\documentclass{ifacconf}

\usepackage{graphicx}      
\usepackage{natbib}        
\usepackage{subcaption}
\usepackage{float}
\usepackage{natbib}        
\usepackage{amsmath}
\usepackage{enumerate}
\usepackage{amssymb}
\usepackage{booktabs}
\usepackage{url}
\begin{document}
\begin{frontmatter}

\title{Efficient Reinforcement Learning from Human Feedback via Bayesian Preference Inference\thanksref{footnoteinfo}} 

\thanks[footnoteinfo]{This work is partially supported by the FAIR project (NextGenerationEU, PNRR-PE-AI, M4C2, Investment 1.3), the 4DDS project (Italian Ministry of Enterprises and Made in Italy, grant F/310097/01-04/X56), and the PRIN PNRR project P2022NB77E (NextGenerationEU, CUP: D53D23016100001). It is also partly supported by the ENFIELD project (Horizon Europe, grant 101120657).}

\author[PoliMI]{Matteo Cercola} 
\author[PoliMI]{Valeria Capretti} 
\author[PoliMI]{Simone Formentin}

\address[PoliMI]{Dipartimento di Elettronica, Informazione e Bioingegneria, Politecnico di Milano, Milano, Italy.(e-mail: author@polimi.it).}

\begin{abstract}             
Learning from human preferences is a cornerstone of aligning machine learning 
models with subjective human judgments. Yet, collecting such preference data is 
often costly and time-consuming, motivating the need for more efficient learning 
paradigms. Two established approaches offer complementary advantages: RLHF 
scales effectively to high-dimensional tasks such as LLM fine-tuning, while PBO 
achieves greater sample efficiency through active querying. We propose a hybrid 
framework that unifies RLHF’s scalability with PBO’s query efficiency by 
integrating an acquisition-driven module into the RLHF pipeline, thereby enabling active and sample-efficient preference gathering. We validate the proposed approach on
two representative domains: (i) high-dimensional preference optimization and 
(ii) LLM fine-tuning. Experimental results demonstrate consistent improvements in both 
sample efficiency and overall performance across these tasks.
\end{abstract}


\end{frontmatter}
\section{Introduction}
Many real-world problems involve objectives that are difficult to specify 
explicitly but easy for humans to recognize intuitively. In such settings, 
humans can reliably state which of two outcomes is preferable, even when they 
cannot articulate a quantitative function that captures this preference. 
For instance, in vehicle suspension tuning, ride comfort is a subjective 
criterion that cannot be directly encoded as a quantitative cost function, 
yet a human can easily express which suspension configuration provides a 
better experience.
This 
gap between human intuition and mathematical formalization has motivated the 
development of frameworks that learn directly from human feedback, allowing 
models to approximate subjective notions of quality without requiring explicit 
reward definitions.\\

In optimization and control, it is often referred to as preference-based 
optimization, where the objective function is implicit and only pairwise 
comparisons are available. Methods such as Preferential Bayesian Optimization 
(PBO) (\cite{Chu2005PreferenceGP,Brochu2010Interactive,gonzalez2017preferential,benavoli2023bayesian})
model latent utilities using Gaussian Processes (GPs) and leverage acquisition 
functions to actively select informative comparisons. While sample-efficient, GP 
surrogates scale poorly in high dimensions.\\
In contrast, within the reinforcement learning community, preference feedback 
has emerged under the paradigm of Reinforcement Learning from Human Feedback 
(RLHF) (\cite{christiano2017deep}). By training a reward model from pairwise human 
judgments and optimizing it through reinforcement learning, RLHF has become a 
key component in aligning large language models (LLMs) with human expectations 
(\cite{ziegler2019fine}). However, standard RLHF is data-hungry, which is 
problematic when annotation budgets are limited.\\
To mitigate this, recent works have begun exploring active preference learning 
strategies that aim to select the most informative comparisons. For instance, 
\cite{sekhari2023contextual} studied query complexity in preference-based 
reinforcement learning, while \cite{ji2024reinforcement} introduced Active 
Proximal Policy Optimization (APPO) for linear dueling bandits with provable 
suboptimality bounds. Other efforts have incorporated query costs directly into 
the reward formulation (\cite{schulze2018active,krueger2020active,tucker2023bandits}), 
highlighting a growing interest in integrating efficiency considerations into 
human-feedback loops.\\
In this work, we introduce a hybrid framework that combines the query efficiency 
of preference-based optimization with the scalability of learning from human feedback, 
thereby bridging the gap between classical PBO and modern RLHF. 
Compared to PBO, the proposed approach leverages neural reward models, as in RLHF, 
and therefore does not suffer from the scalability limitations that arise when using 
Gaussian Process surrogates in high-dimensional settings. 
Relative to standard RLHF, it introduces a Bayesian acquisition mechanism that 
actively selects informative preference queries, substantially improving data 
efficiency under constrained annotation budgets. 
Unlike recent efforts to enhance RLHF through ensemble- or logit-based uncertainty 
approximations, our method employs a Laplace-based Bayesian estimation that 
provides a theoretically grounded and computationally lightweight measure of 
model uncertainty, which can be seamlessly integrated into existing RLHF training 
pipelines. 
The resulting framework, referred to as \textbf{Bayesian RLHF}, operates seamlessly 
across domains—from fine-tuning large language models to optimizing high-dimensional 
continuous functions, and consistently outperforms RLHF and PBO in both 
sample efficiency and accuracy.
The main contributions of this work are as follows:
\begin{itemize}
    \item We incorporate a \textbf{Laplace-based Bayesian uncertainty estimation} within the RLHF pipeline, providing a principled means of uncertainty quantification without relying on costly ensembles or heuristic logit-based approximations.
    \item Building upon Dueling Thompson Sampling, we introduce a \textbf{mixing acquisition strategy} that balances exploration and exploitation through a fixed mixing coefficient, enhancing sample efficiency while preserving scalability to high-dimensional neural settings.
    \item We validate the proposed framework on two complementary domains, high-dimensional numerical optimization and LLM fine-tuning from human preferences, demonstrating consistent improvements in both \textbf{sample efficiency} and \textbf{final performance}.
\end{itemize}

\section{PRELIMINARIES}

\subsection{Reinforcement Learning from Human Feedback (RLHF)}
Reinforcement Learning from Human Feedback (RLHF) (\cite{christiano2017deep}) is a 
framework designed to align policy behavior with human preferences through an 
iterative loop involving data collection, reward modeling, and policy 
optimization.  
In its standard formulation, RLHF starts from a pretrained policy $\pi_{\theta}$, 
typically a neural network or a language model, that generates candidate outputs 
for a given input. Human evaluators then provide pairwise comparisons between 
outputs, indicating which is preferred. These comparisons are used to train a 
reward model $r_{\phi}(x, y)$, typically a neural network, that predicts the 
relative preference between two outputs $y_1$ and $y_2$ conditioned on an input 
$x$:
\begin{equation}
    P(y_1 \succ y_2 \mid x) = \sigma\big(r_{\phi}(x, y_1) - r_{\phi}(x, y_2)\big),
\end{equation}
where $\sigma(\cdot)$ denotes the logistic function.\\
The trained reward model, then guides policy optimization, typically via PPO 
(\cite{schulman2017proximal}). \\
While RLHF has demonstrated remarkable success in aligning large language models 
with human intent, its data collection process remains a critical bottleneck. In 
the original formulation, query selection relies on estimating uncertainty in 
the reward model by maintaining an ensemble of predictors and sampling those 
comparisons that yield the highest disagreement among ensemble members. Although 
this approach provides a rough measure of uncertainty, it represents only a 
crude approximation of optimal information gain. The authors themselves noted 
that, in some tasks, this strategy even degraded performance.

\subsection{Preferential Bayesian Optimization (PBO)}
Preferential Bayesian Optimization (PBO) (\cite{Chu2005PreferenceGP,Brochu2010Interactive,gonzalez2017preferential,benavoli2023bayesian})
extends the principles of Bayesian Optimization (BO) to settings where the 
objective function is unknown and can only be queried through pairwise 
preferences rather than scalar evaluations.  
Given two candidate solutions $\mathbf{x}_1$ and $\mathbf{x}_2$, the human 
provides a preference $\mathbf{x}_1 \succ \mathbf{x}_2$ if $\mathbf{x}_1$ is 
judged better. These comparisons are modeled by a latent utility function 
$f(\mathbf{x})$, endowed with a Gaussian Process (GP) prior. The likelihood of a 
preference is modeled via a Bernoulli distribution:
\begin{equation}
    P(\mathbf{x}_1 \succ \mathbf{x}_2) = \Phi\!\left(\frac{f(\mathbf{x}_1) - 
    f(\mathbf{x}_2)}{\sqrt{2}\sigma}\right),
\end{equation}
where $\Phi$ is the cumulative Gaussian function and $\sigma$ represents 
observation noise.\\
At each iteration, PBO uses the GP posterior to compute an acquisition function 
$a(\mathbf{x})$ that quantifies the expected utility of querying a given point. 
\begin{equation}
    (\mathbf{x}_i, \mathbf{x}_j) = \arg\max_{\mathbf{x}_i, \mathbf{x}_j} 
    a(\mathbf{x}_i, \mathbf{x}_j)
\end{equation}
where $a$ aims to maximize expected information gain about $f$, or equivalently, 
reduce posterior uncertainty. This active querying mechanism enables PBO to 
focus human feedback on the most informative comparisons, dramatically improving 
sample efficiency. However, the reliance on Gaussian Process priors causes the 
computational cost and memory footprint to scale poorly with both the number of 
dimensions and the number of observations, making it less suitable for 
large-scale or high-dimensional problems such as LLM fine-tuning.

\subsection{Laplace Approximation}
The \emph{Laplace Approximation} (LA) is a classical Bayesian technique that 
approximates the posterior distribution of a model’s parameters without requiring 
architectural changes or ensembles of networks. Conceptually, LA constructs a 
local Gaussian approximation of the posterior distribution centered at the most 
probable parameters of the model.\\
From a Bayesian perspective, minimizing a regularized loss function can be 
interpreted as finding the \emph{maximum-a-posteriori} (MAP) estimate:
\begin{equation}
    \theta_{\text{MAP}} = \arg \min_{\theta} \mathcal{L}(\mathcal{D}; \theta),
\end{equation}
where $\mathcal{L}(\mathcal{D}; \theta)$ denotes the loss computed over the 
dataset $\mathcal{D}$.\\
The Laplace approximation replaces the original loss with its second-order 
Taylor expansion around $\theta_{\text{MAP}}$:
\begin{equation}
\begin{split}
\mathcal{L}(\mathcal{D}; \theta) 
&\approx \mathcal{L}(\mathcal{D}; \theta_{\text{MAP}}) 
+ \frac{1}{2} (\theta - \theta_{\text{MAP}})^{\top} 
H (\theta - \theta_{\text{MAP}}),\\
H &= \nabla^{2}_{\theta} \mathcal{L}(\mathcal{D}; \theta)\big|_{\theta = \theta_{\text{MAP}}}.
\end{split}
\label{eq:laplace_taylor}
\end{equation}\\
Since the expansion is performed around the minimum point $\theta_{\text{MAP}}$, 
the first-order term vanishes.  
After standard algebraic manipulations, it can be shown that the posterior 
distribution can be approximated as:
\begin{equation}
    P(\theta \mid \mathcal{D}) \approx 
    \mathcal{N}\big(\theta_{\text{MAP}}, H^{-1}\big),
    \label{eq:laplace_posterior}
\end{equation}
which implies that, after training, the posterior distribution of the parameters 
is Gaussian centered at the MAP estimate and with covariance given by the 
inverse Hessian.\\
Thus, a deterministic network gains a local Gaussian posterior over parameters.

\section{Main Contribution}
\begin{figure}[h!]
    \centering
    \includegraphics[width=\linewidth,trim=20 40 20 100]{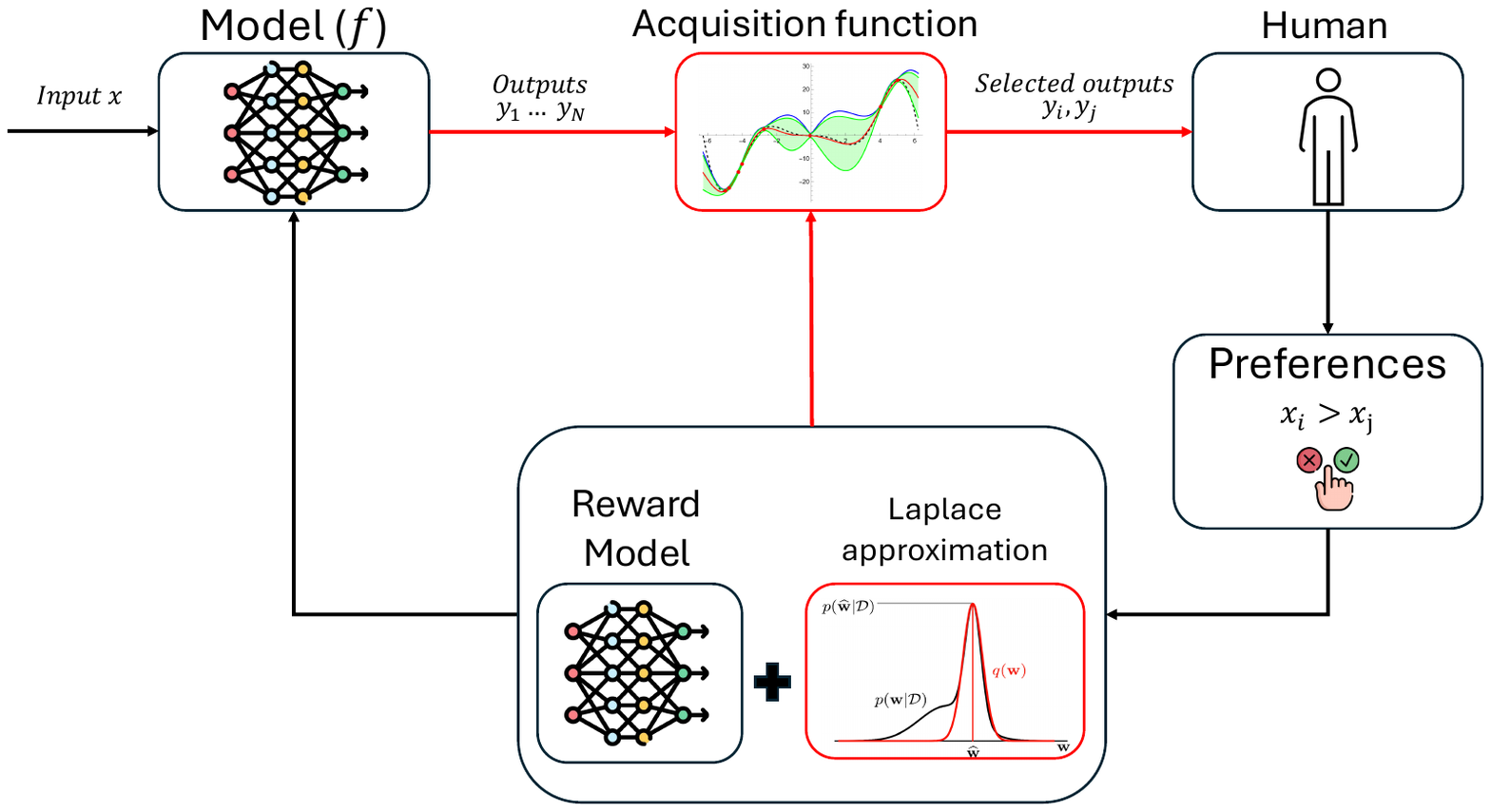}
    \caption{Overview of the proposed Bayesian RLHF framework, integrating 
    Laplace-based uncertainty estimation in the reward model and an acquisition 
    function for efficient preference querying. Novel components relative to 
    standard RLHF are highlighted in red.}
    \label{fig:B_RLHF_framework}
\end{figure}

This section introduces the proposed hybrid framework, termed Bayesian RLHF. As 
illustrated in Fig.~\ref{fig:B_RLHF_framework}, our approach introduces two key 
additions to the classical RLHF loop: (i) a \textit{Laplace-based uncertainty 
estimation} in the reward model, and (ii) an \textit{acquisition function} that 
exploits this uncertainty to actively guide human queries.  
Both extensions are lightweight and compatible with existing RLHF frameworks, 
such as Hugging Face \texttt{trl} (\cite{vonwerra2022trl}). The Laplace 
approximation is applied \textit{post hoc} to a pre-trained reward model, 
without requiring any modification to the training pipeline or architecture.

\subsection{Laplace-Based Reward Model}
In classical RLHF, the reward model $r_\phi$ is trained via maximum likelihood 
on pairwise preferences $(y_1, y_2)$ to approximate the human-judged preference 
probability:
\begin{equation}
    P(y_1 \succ y_2 \mid x) = \sigma\big(r_\phi(x, y_1) - r_\phi(x, y_2)\big),
\end{equation}\\
where $\sigma(\cdot)$ denotes the logistic link.  
This formulation corresponds to a \emph{binary classification task} with 
discrete feedback, where the model learns to predict the preferred output 
between two alternatives.  
However, standard training yields only a point estimate $\phi_{\text{MAP}}$ of 
the reward parameters, providing no notion of uncertainty about the learned 
preferences. To endow the reward model with calibrated uncertainty estimates, we 
apply the Laplace Approximation (\cite{daxberger2021laplace}). First, the reward 
model is trained to obtain the $\phi_{\text{MAP}}$. The second step consists of 
computing the Hessian matrix $H$, which captures the local curvature of the loss 
around $\phi_{\text{MAP}}$. Nevertheless, using LA within an iterative framework 
such as RLHF introduces several challenges:  
(1) Computing the Hessian requires a full pass over the fine-tuning dataset, 
which continually grows during training.  
(2) Storing the Hessian is infeasible, as its size scales quadratically with the 
number of parameters (which are typically very large in the context of LLMs).  
(3) Evaluating the Hessian can be computationally prohibitive, and the resulting 
matrix may be indefinite. \\ 
To overcome these limitations, instead of resorting to large-scale 
approximations (e.g., Fisher information or diagonalized Hessians), we adopt a 
more practical strategy: compute the exact Hessian only for a small subset of 
parameters, typically those in the final layer of the network. This approach, 
commonly referred to as the last-layer Laplace approximation, allows the method 
to scale efficiently to large models. In the context of RLHF with an LLM-based 
reward model, this corresponds to the compact classification head (approximately 
512 parameters) attached to the frozen language model backbone.\\

\subsection{Acquisition-Driven Query Selection}
While the Laplace approximation endows the reward model with calibrated 
uncertainty estimates, the next step is to exploit this uncertainty to actively 
select informative preference queries. Therefore, we introduce an 
\emph{acquisition-driven query selection} mechanism inspired by PBO.  
Our strategy is based on \textit{Dueling Thompson Sampling} (DTS), an acquisition 
function that explicitly balances exploration and exploitation.  
At each iteration, a set of candidate responses $\{y_i\}_{i=1}^M$ is evaluated 
by the Bayesian reward model, producing both mean and variance estimates of 
their scores.  
The \emph{best} candidate $y_\text{best}$ is chosen as the one with the highest 
sampled utility from the posterior distribution, and a \emph{rival} candidate 
$y_\text{rival}$ is then selected according to one of two complementary modes.

\medskip
\textbf{Sparring Mode — Exploitation.}
In this mode, the rival is sampled stochastically among strong candidates to 
favor duels likely to refine the current preference boundary.  
Let $s(y_i)$ denote the predicted win score for candidate $y_i$; the rival is 
drawn from a softmax distribution controlled by a temperature $T$:
\begin{equation}
    y'_{\text{rival, spar}} \sim \text{Softmax}\left(\frac{s(y_i)}{T}\right),
\end{equation}
where lower $T$ values emphasize exploitation by amplifying score differences, 
while higher $T$ promotes broader sampling.  

\medskip
\textbf{MaxVar Mode — Exploration.}
Conversely, in the \emph{MaxVar} mode the rival is chosen as the candidate 
inducing the highest predictive uncertainty with respect to the current best:
\begin{equation}
    y'_{\text{rival, maxvar}} = \arg\max_{y_i \in \mathcal{Y}} 
    \text{Var}\big[p(y_\text{best} \succ y_i)\big],
\end{equation}
where $\text{Var}[p(\cdot)]$ denotes the posterior variance of the win 
probability estimated via the Laplace-based reward model.  
This encourages duels that are maximally informative for refining the model’s 
belief, thereby encouraging exploration of uncertain regions in the preference 
space.

\medskip
\textbf{Mixed Strategy — Balancing Exploration and Exploitation.}
To flexibly navigate between exploration and exploitation, we introduce a 
\emph{convex combination} of the two modes.  
The acquisition score $J_\alpha(y')$ for a candidate $y'$ is defined as:
\begin{equation}
J_{\alpha}(y') =
\alpha \, \frac{S_{\mathrm{spar}}(y') - 
\mathbb{E}[S_{\mathrm{spar}}]}{\mathrm{Std}[S_{\mathrm{spar}}]}
+ (1 - \alpha) \, \frac{S_{\mathrm{var}}(y') - 
\mathbb{E}[S_{\mathrm{var}}]}{\mathrm{Std}[S_{\mathrm{var}}]}.
\label{eq:score_combination}
\end{equation}
where $\alpha \in [0,1]$ controls the trade-off between the exploitation-
oriented Sparring score and the exploration-oriented MaxVar score.  
For $\alpha = 1$, the system prioritizes refinement of strong candidates 
(Sparring), while $\alpha = 0$ targets uncertain duels to expand the learned 
preference space (MaxVar).

\medskip
The proposed acquisition-driven query selection transforms RLHF from a passive 
preference learner into an \emph{active querying framework}.  

\subsection{Theoretical Motivation.}\label{sec:theoretical_analysis}
Before presenting the experimental results, it is worth highlighting that the 
proposed hybrid Bayesian RLHF method inherits theoretical advantages over 
classical PBO in high-dimensional settings. Compared to Gaussian-process-based 
PBO, the proposed Bayesian RLHF framework achieves:
\begin{enumerate}
    \item \textbf{Improved scalability.} The cumulative regret of PBO is governed 
    by the information gain term, which, as demonstrated by~\cite{srinivas2012information}, 
    grows rapidly with the dimensionality of the input space. This results in an 
    exponential increase in the number of queries required to reach comparable 
    performance as the problem dimension increases, making GP-based methods 
    impractical in high-dimensional spaces.

    \item \textbf{Reduced computational complexity.}  
    In PBO, posterior updates over the latent preference function require approximate 
    inference.  
    When the Laplace approximation is employed, the dominant computational cost 
    arises from cubic matrix operations, $\mathcal{O}(T^3)$, with respect to the 
    number of queries $T$.  
    In contrast, the proposed Bayesian RLHF method applies the Laplace approximation 
    only to the final layer of the neural reward model, which typically contains a 
    small and fixed number of parameters. 
\end{enumerate}
These theoretical considerations suggest that the proposed hybrid Bayesian RLHF 
method maintains the expressiveness and scalability of neural models while 
retaining the uncertainty quantification benefits of Laplace-based Bayesian 
inference.

\section{Experimental Results}
This section presents a numerical evaluation of the proposed Bayesian RLHF 
(B-RLHF) on (i) numerical optimization against PBO and (ii) LLM fine-tuning 
compared to RLHF.

\subsection{High-Dimensional Preference Optimization}
\begin{figure*}[t]
    \centering
    \begin{subfigure}[b]{0.48\textwidth}
        \centering
        \includegraphics[width=\linewidth]{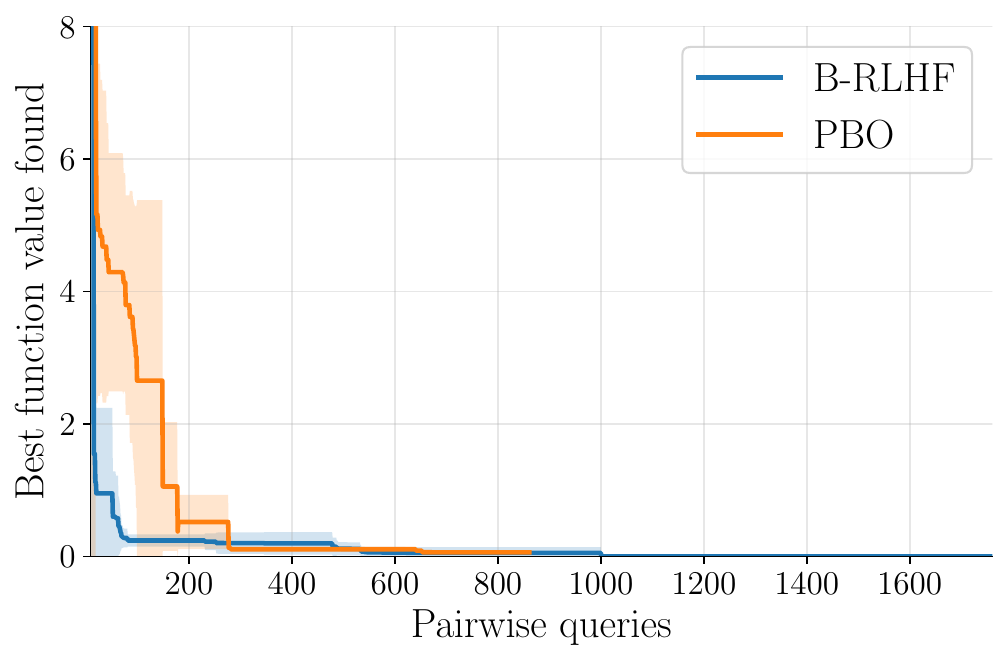}
        \caption{2D Rosenbrock optimization.}
        \label{fig:2d_numerical_eval}
    \end{subfigure}
    \hfill
    \begin{subfigure}[b]{0.48\textwidth}
        \centering
        \includegraphics[width=\linewidth]{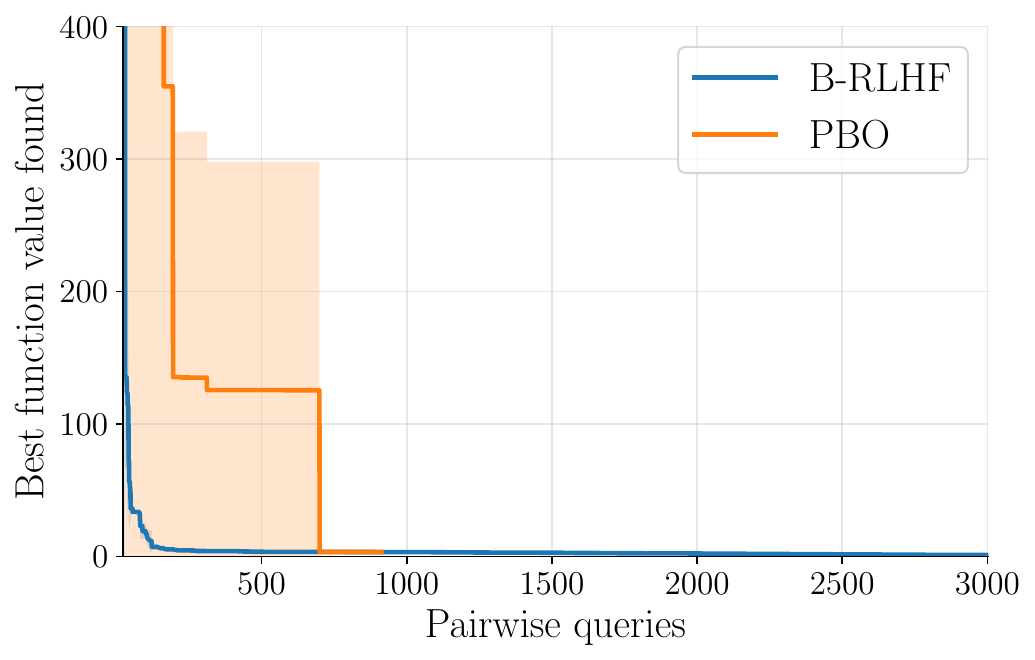}
        \caption{5D Rosenbrock optimization.}
        \label{fig:5D_numerical_eval}
    \end{subfigure}
       \caption{Comparison between Bayesian RLHF (B-RLHF) in blue and baseline PBO 
       in orange on the Rosenbrock problem. Solid lines indicate the mean response, 
       and the shaded bands represent $\pm$ one standard deviation over 5 Monte Carlo 
       runs.}
    \label{fig:2d_5D_numerical_eval}
\end{figure*}

\begin{figure*}[t]
    \centering
    \begin{subfigure}[b]{0.48\textwidth}
        \centering
        \includegraphics[width=\linewidth]{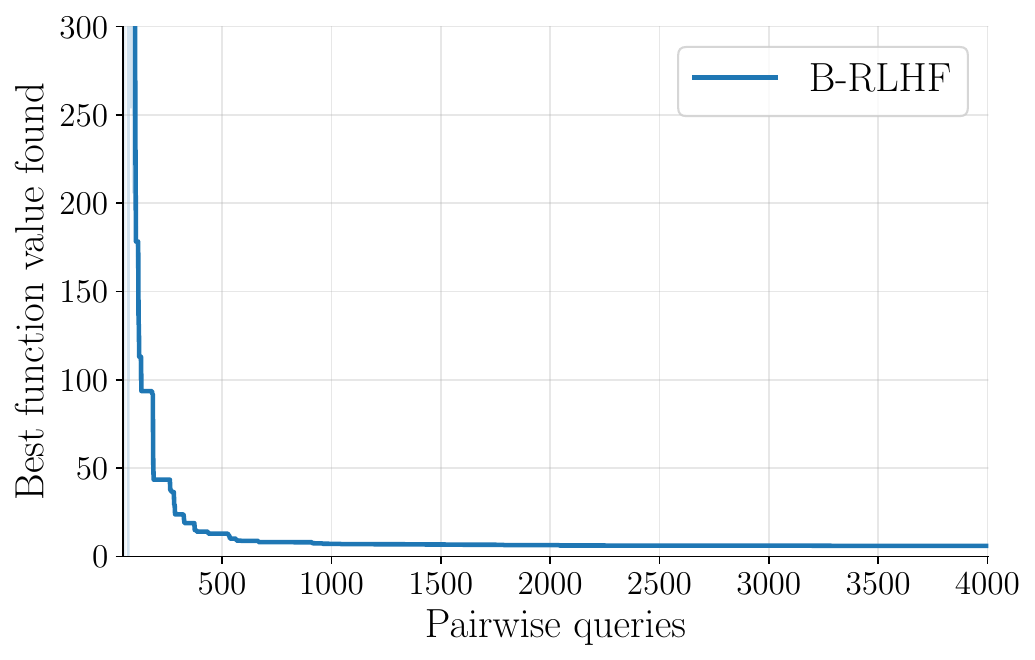}
        \caption{10D Rosenbrock optimization.}
        \label{fig:10d_numerical_eval}
    \end{subfigure}
    \hfill
    \begin{subfigure}[b]{0.48\textwidth}
        \centering
        \includegraphics[width=\linewidth]{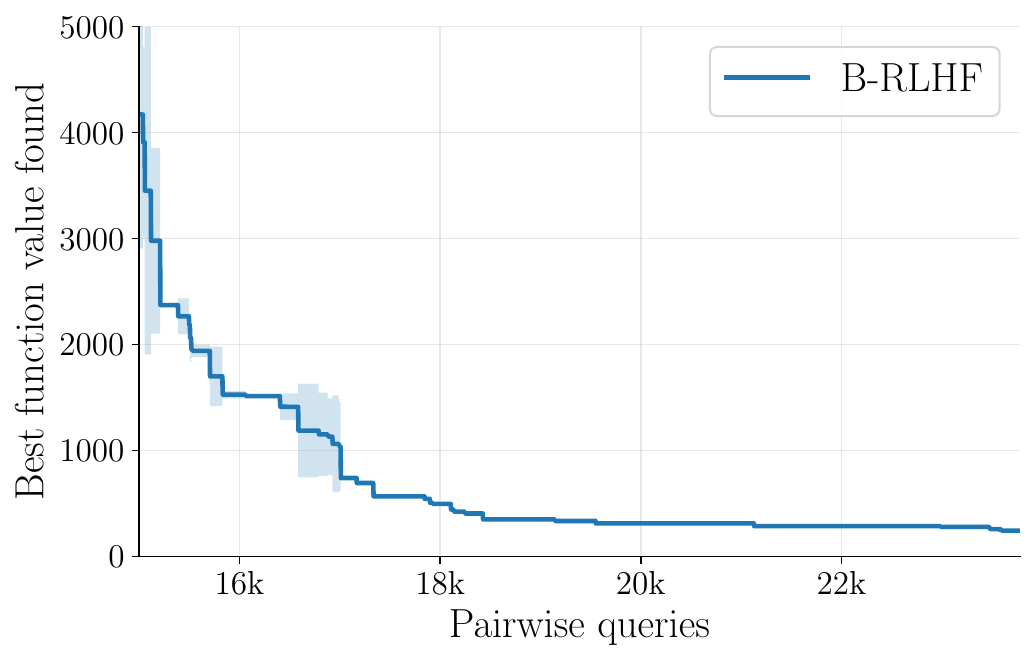}
        \caption{50D Rosenbrock optimization.}
        \label{fig:50D_numerical_eval}
    \end{subfigure}
       \caption{Best value of the latent function on the Rosenbrock optimization 
       problem, achieved by our algorithm Bayesian RLHF (B-RLHF). Solid lines 
       indicate the mean response, and the shaded bands represent $\pm$ one standard 
       deviation over 3 Monte Carlo runs.}
    \label{fig:10d_50D_numerical_eval}
\end{figure*}

\begin{figure}[h!]
    \centering
    \includegraphics[width=0.9\linewidth,trim={0pt 0pt 0pt 0pt},clip]{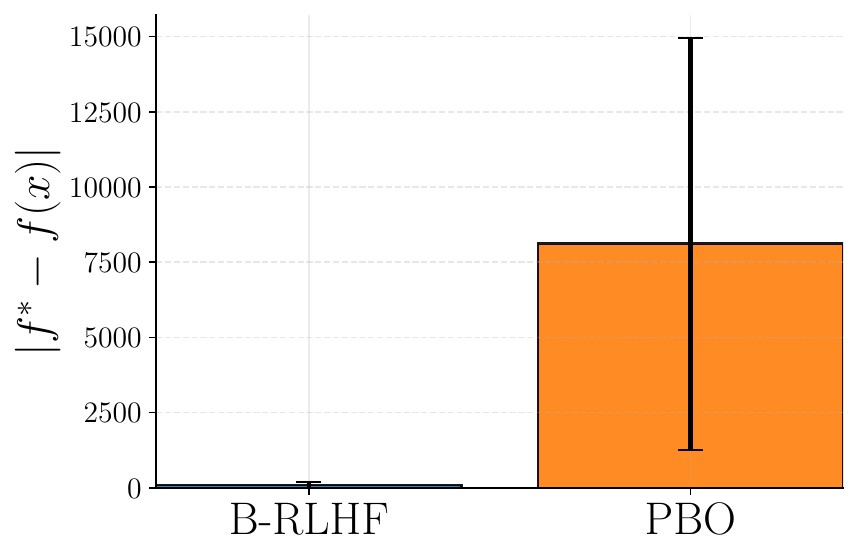}
    \caption{Mean and standard deviation of the final optimization error across 
    3 independent runs for B-RLHF and PBO on the 10D Rosenbrock function with a 
    budget of 4000 queries, a 10-hour time limit.}
    \label{fig:10D_bar}
\end{figure}

The optimization task is based on the $d$-dimensional Rosenbrock function, a 
nonlinear benchmark characterized by a narrow, curved valley that makes both 
exploration and convergence particularly challenging.\\
For this experiment, we implemented the two algorithms as follows: 
\begin{itemize}
    \item \textbf{Bayesian RLHF:} both the policy and reward models are implemented 
    as neural networks. The reward model is trained from pairwise comparisons using 
    a Bradley–Terry logistic loss and consists of a deep multilayer perceptron (MLP) 
    backbone with a linear head. A Laplace approximation is applied to the last 
    layer.  The acquisition function parameter $\alpha$ (see Eq.~\ref{eq:score_combination}) 
    is set equal to 0.5 to balance exploration and exploitation.
    \item \textbf{PBO:} we implemented a Matérn kernel PBO baseline that employs 
    Laplace-based uncertainty estimation.
\end{itemize}
All experiments were executed on identical hardware (Intel Xeon 2.00 GHz CPU, 13 
GB RAM, NVIDIA Tesla P100 GPU with 16 GB VRAM) to ensure fairness. Uniform 
stopping criteria were enforced: (i) a computational budget proportional to the 
problem dimensionality (larger dimensions implying larger budgets), and (ii) a 
fixed wall-clock time limit of 10 hours.\\
Figure \ref{fig:2d_numerical_eval} compares our Bayesian RLHF method with PBO on 
the 2D Rosenbrock problem across 5 Monte Carlo runs. Our approach achieves 
faster convergence, and reaches a 44\% lower error on average at the final 
iteration. The Bayesian RLHF fully exploited the available queries budget within 
the time limit, whereas the baseline PBO required significantly longer runtime, 
completing on average only 850 preference queries within the same time limit, 
which is consistent with its theoretical cubic complexity in the number of 
queries as discussed in Section \ref{sec:theoretical_analysis}.\\
Similar trends are observed in the 5D case (Figure \ref{fig:5D_numerical_eval}), 
where Bayesian RLHF exhibits superior convergence speed and solution quality, 
reaching the final error achieved by PBO 200 queries earlier while using only 
20\% of the full query budget.\\
Figure \ref{fig:10d_50D_numerical_eval} extends the comparison to high-dimensional 
settings (10D and 50D Rosenbrock). In the 10D case, PBO failed to complete the 
optimization due to memory exhaustion after 650 queries, while our method 
successfully converged within the full query budget of 4000. Figure~\ref{fig:10D_bar} 
reports the final absolute error at termination for both methods, further 
underscoring the superiority of our approach. At 50 dimensions, PBO becomes 
computationally infeasible, whereas Bayesian RLHF, Figure \ref{fig:50D_numerical_eval}, 
continues to make progress throughout the 10-hour window, demonstrating a 
consistent convergence trend despite not fully exhausting the budget.

\begin{figure}[h!]
    \centering
    \includegraphics[width=\linewidth,trim={0pt 0pt 0pt 0pt},clip]{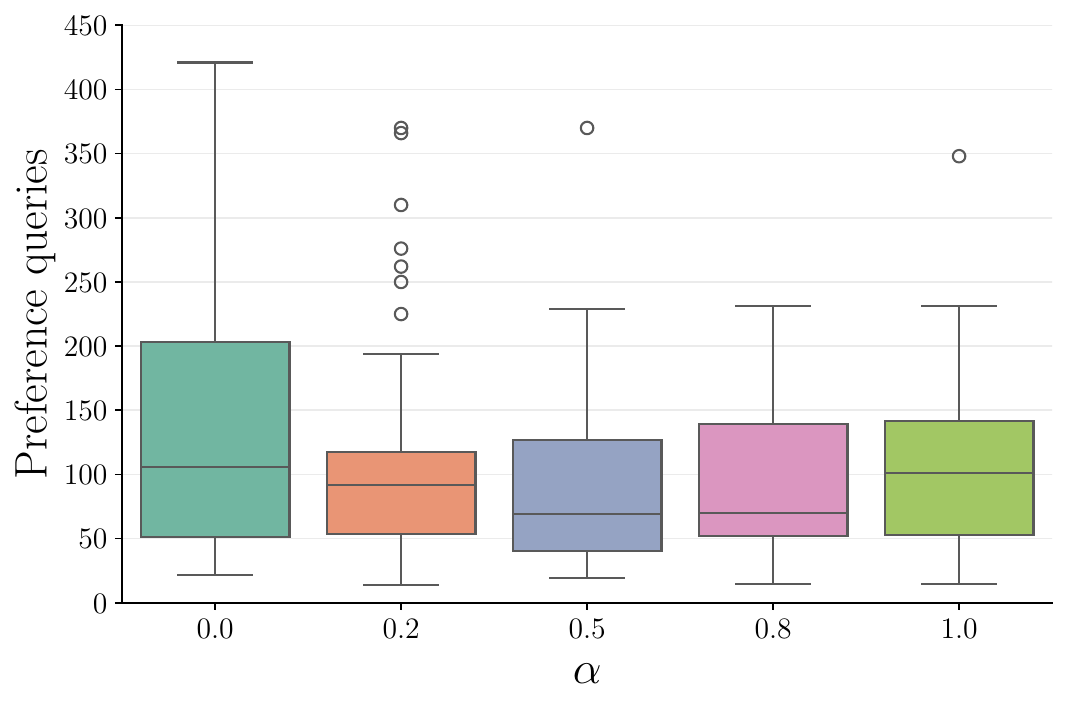}
    \caption{Sensitivity analysis of the $\alpha$ exploration–exploitation 
    parameter, averaged over 38 Monte Carlo runs.}
    \label{fig:Alpha_sweep}
\end{figure}

Finally, in this numerical optimization setting, we performed a sensitivity 
analysis of the $\alpha$ parameter controlling the exploration–exploitation 
balance in our acquisition function. The results are shown in Figure 
\ref{fig:Alpha_sweep}. We performed 38 Monte Carlo runs on the 2D Rosenbrock 
function, varying $\alpha$ from 0 (pure max-variance exploration) to 1 (pure 
exploitation). Results indicate that intermediate values of $\alpha$ yield the 
most sample-efficient optimization, requiring the fewest preference queries to 
reach the optimum, highlighting the benefit of a mixed exploration–exploitation 
strategy. In this experiment, the best-performing value of $\alpha$ was $0.5$, which achieved the lowest median and interquartile range of queries required to reach the optimum.

\subsection{LLM Fine-Tuning}
\begin{figure}[h!]
    \centering
    \includegraphics[width=\linewidth,clip]{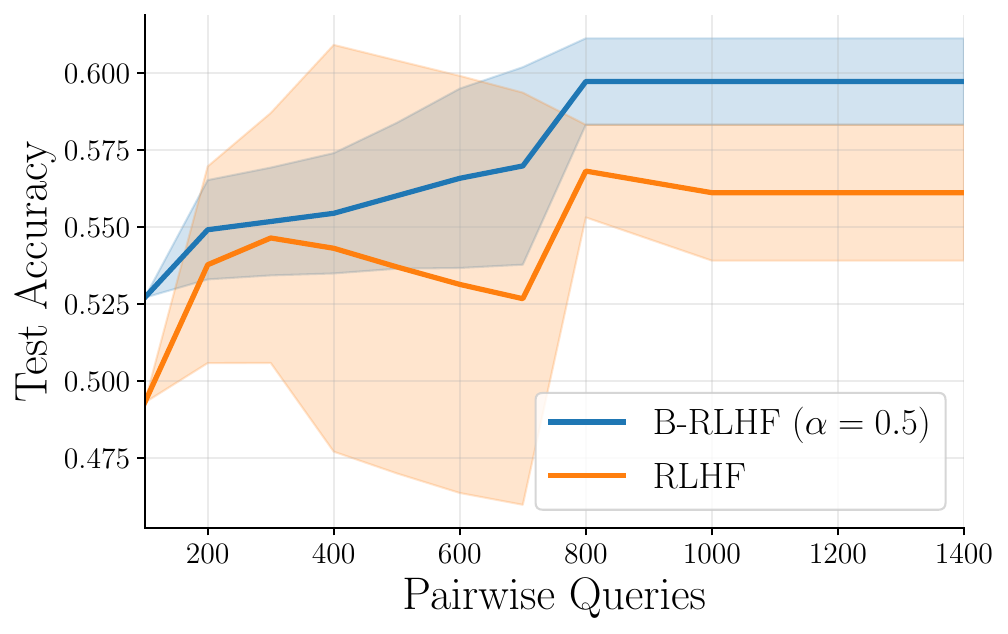}
    \caption{Comparison between Bayesian RLHF (B-RLHF) with $\alpha = 0.5$, which 
    achieved the best overall accuracy across all tested $\alpha$ values, and the 
    baseline RLHF on the LLM fine-tuning problem. Solid lines denote the mean 
    performance, and shaded regions indicate $\pm$ one standard deviation over three 
    Monte Carlo runs.}
    \label{fig:LLM_results_MC}
\end{figure}

To further validate our approach, we extend the analysis to a language-model 
fine-tuning scenario based on human preferences. We derive both the base and 
reward networks from the \texttt{Pythia-70M}\footnote{\url{https://huggingface.co/EleutherAI/pythia-70m}} 
architecture and trained on the publicly available \texttt{Dahoas/rm-hh-rlhf}\footnote{\url{https://huggingface.co/datasets/Dahoas/rm-hh-rlhf}} 
dataset. Each record contains a user prompt and two candidate assistant responses, 
where the one labeled as \textit{“chosen”} reflects the human-preferred completion 
over the \textit{“rejected”} one. 
The dataset comprises approximately 112K training 
and 12.5K test samples within the Helpful \& Harmless (HH) dialogue preference 
framework. For each training iteration, prompts are sampled from the static \texttt{Dahoas/rm-hh-rlhf} dataset, while the corresponding candidate responses are generated online by the policy model.
In the baseline RLHF configuration, two responses are sampled uniformly from the model outputs, following the standard approach used in \cite{ziegler2019fine}.
In contrast, our Bayesian RLHF implementation employs an uncertainty-based acquisition function to select the response pair expected to be most informative according to the reward model. This setup reproduces an online active preference selection process, where querying decisions depend dynamically on model uncertainty rather than uniform sampling. Following established RLHF literature (\cite{eisenstein2023helping,coste2023reward,shen2024improving}), 
we employed \textit{PairRM}(\cite{jiang2023llm}) as a proxy human annotator, a reward 
model trained on the large-scale \texttt{UltraFeedback}\footnote{\url{https://huggingface.co/datasets/openbmb/UltraFeedback}} 
dataset, which encompasses the HH dataset and is renowned for its diversity and 
high-quality annotations. All experiments were conducted on identical hardware 
equipped with an NVIDIA A100 GPU featuring 32~GB of VRAM\\
Since the reward model is the principal component modified in our framework and 
serves as a proxy for human feedback during policy optimization, we focus our 
evaluation on its \textbf{predictive accuracy}. The reward model was fine-tuned 
with the last two layers unfrozen, corresponding to approximately 6 million 
parameters (16\% of the total), using a learning rate of $6\times10^{-4}$. The 
model parameters were updated every 200 preference queries. Results in table~\ref{tab:agg_results} 
compares our Bayesian RLHF method with the baseline RLHF on the LLM fine-tuning 
task, averaged over three Monte Carlo runs. In this setting, we used 1{,}400 
pairwise preferences for training and 500 unseen prompts for testing. All 
configurations of our method (for all tested values of $\alpha$) achieved higher 
final accuracy than the baseline RLHF. In particular, Figure~\ref{fig:LLM_results_MC} 
compares the best-performing configuration ($\alpha = 0.5$) against the baseline, 
showing a \textbf{6\% improvement in mean accuracy}. An additional experiment 
performed with an increased preference budget is summarized in Table~\ref{tab:single_result}. 
The best-performing configuration ($\alpha = 1$) achieved a \textbf{14\% improvement} 
over standard RLHF. Interestingly, with the larger preference budget, the optimal configuration shifted from $\alpha = 0.5$ to $\alpha = 1$.
This behavior can be explained by the reduced uncertainty of the reward model at higher data volumes: once the model achieves sufficient predictive accuracy, exploration offers diminishing returns, and a purely exploitative strategy becomes more effective in driving convergence. \\
Across both experiments, our Bayesian RLHF method consistently outperformed 
standard RLHF. Notably, even with a limited number of pairwise queries (at most 
3.1\% of the available dataset), the method demonstrated robust improvements. We 
deliberately constrained the query budget for two reasons:
\begin{enumerate}
\item To emulate realistic human-in-the-loop settings, where preference 
collection is expensive.
\item To shorten training time and emphasize sample efficiency.
\end{enumerate}

\begin{table}[ht]
\centering
\caption{Test-set accuracy for Bayesian RLHF (B-RLHF) and baseline RLHF on the 
LLM fine-tuning task, averaged over three Monte Carlo runs. Each run uses 1{,}400 
training and 500 testing preference queries. Reported values correspond to the 
accuracy at the final iteration, with mean and standard deviation computed across 
3 Monte Carlo runs.}
\label{tab:agg_results}
\begin{tabular}{lccc}
\toprule
\textbf{Method} &\textbf{$\alpha$}& \textbf{Mean} & \textbf{Std.} \\
\midrule
B-RLHF &0       & 0.596 & 0.001 \\
B-RLHF &0.2 & 0.585 & 0.008 \\
B-RLHF &0.5 & \textbf{0.597} & 0.014\\
B-RLHF &0.8 & 0.577 & 0.0261\\
B-RLHF &1          & 0.593 & 0.001\\
RLHF  &-          & 0.561 & 0.022 \\
\bottomrule
\end{tabular}
\end{table}

\begin{table}[ht]
\centering
\caption{Test-set accuracy of the Bayesian RLHF (B-RLHF) and baseline RLHF 
models on the LLM fine-tuning task, trained with an extended dataset comprising 
3{,}500 training and 1{,}000 testing preference queries. Results are reported 
from a single run and correspond to the accuracy at the final iteration.}
\label{tab:single_result}
\begin{tabular}{lccc}
\toprule
\textbf{Method} &\textbf{$\alpha$}& \textbf{Accuracy} \\
\midrule
B-RLHF &0       & 0.587 \\
B-RLHF &0.2 & 0.589\\
B-RLHF &0.5 & 0.616\\
B-RLHF &0.8 & 0.615 \\
B-RLHF &1          & \textbf{0.635}\\
RLHF  &-          &  0.549  \\
\bottomrule
\end{tabular}
\end{table}

\section{Conclusions}
This paper introduced \emph{Bayesian RLHF}, a hybrid framework that combines the 
sample-efficiency benefits of preference-based optimization with the scalability of 
learning from human feedback. 
Across both numerical optimization and language-model fine-tuning tasks, the proposed 
approach demonstrated faster convergence and higher test accuracy under tight query 
budgets, supporting our central claim on data efficiency. 

The present study is limited by the evaluation scope, which primarily assessed 
reward-model accuracy using a proxy annotator. Extending the analysis to full 
end-to-end policy optimization with human raters represents an important direction 
for future work. 

Future research will also investigate an adaptive exploration–exploitation 
trade-off within the acquisition mechanism, as well as the use of the proposed 
uncertainty quantification in the policy optimization step to further enhance 
stability and performance. 

\bibliography{ifacconf}    

\begin{thebibliography}{19}
\providecommand{\natexlab}[1]{#1}
\providecommand{\url}[1]{\texttt{#1}}
\providecommand{\urlprefix}{URL }
\expandafter\ifx\csname urlstyle\endcsname\relax
  \providecommand{\doi}[1]{doi:\discretionary{}{}{}#1}\else
  \providecommand{\doi}{doi:\discretionary{}{}{}\begingroup \urlstyle{rm}\Url}\fi

\bibitem[{Benavoli et~al.(2023)Benavoli, Azzimonti, and Piga}]{benavoli2023bayesian}
Benavoli, A., Azzimonti, D., and Piga, D. (2023).
\newblock Bayesian optimization for choice data.
\newblock In \emph{Proceedings of the Companion Conference on Genetic and Evolutionary Computation}, 2272--2279.

\bibitem[{Brochu(2010)}]{Brochu2010Interactive}
Brochu, E. (2010).
\newblock \emph{Interactive Bayesian optimization: learning user preferences for graphics and animation}.
\newblock Ph.D. thesis, University of British Columbia.

\bibitem[{Christiano et~al.(2017)Christiano, Leike, Brown, Martic, Legg, and Amodei}]{christiano2017deep}
Christiano, P.F., Leike, J., Brown, T., Martic, M., Legg, S., and Amodei, D. (2017).
\newblock Deep reinforcement learning from human preferences.
\newblock \emph{Advances in neural information processing systems}, 30.

\bibitem[{Chu and Ghahramani(2005)}]{Chu2005PreferenceGP}
Chu, W. and Ghahramani, Z. (2005).
\newblock Preference learning with gaussian processes.
\newblock In \emph{Proceedings of the 22nd International Conference on Machine Learning (ICML)}, 137--144. Pittsburgh, PA, USA.

\bibitem[{Coste et~al.(2023)Coste, Anwar, Kirk, and Krueger}]{coste2023reward}
Coste, T., Anwar, U., Kirk, R., and Krueger, D. (2023).
\newblock Reward model ensembles help mitigate overoptimization.
\newblock \emph{arXiv preprint arXiv:2310.02743}.

\bibitem[{Daxberger et~al.(2021)Daxberger, Kristiadi, Immer, Eschenhagen, Bauer, and Hennig}]{daxberger2021laplace}
Daxberger, E., Kristiadi, A., Immer, A., Eschenhagen, R., Bauer, M., and Hennig, P. (2021).
\newblock Laplace redux-effortless bayesian deep learning.
\newblock \emph{Advances in neural information processing systems}, 34, 20089--20103.

\bibitem[{Eisenstein et~al.(2023)Eisenstein, Nagpal, Agarwal, Beirami, D'Amour, Dvijotham, Fisch, Heller, Pfohl, Ramachandran et~al.}]{eisenstein2023helping}
Eisenstein, J., Nagpal, C., Agarwal, A., Beirami, A., D'Amour, A., Dvijotham, D., Fisch, A., Heller, K., Pfohl, S., Ramachandran, D., et~al. (2023).
\newblock Helping or herding? reward model ensembles mitigate but do not eliminate reward hacking.
\newblock \emph{arXiv preprint arXiv:2312.09244}.

\bibitem[{Gonz{\'a}lez et~al.(2017)Gonz{\'a}lez, Dai, Damianou, and Lawrence}]{gonzalez2017preferential}
Gonz{\'a}lez, J., Dai, Z., Damianou, A., and Lawrence, N.D. (2017).
\newblock Preferential bayesian optimization.
\newblock In \emph{International Conference on Machine Learning}, 1282--1291. PMLR.

\bibitem[{Ji et~al.(2024)Ji, He, and Gu}]{ji2024reinforcement}
Ji, K., He, J., and Gu, Q. (2024).
\newblock Reinforcement learning from human feedback with active queries.
\newblock \emph{arXiv preprint arXiv:2402.09401}.

\bibitem[{Jiang et~al.(2023)Jiang, Ren, and Lin}]{jiang2023llm}
Jiang, D., Ren, X., and Lin, B.Y. (2023).
\newblock Llm-blender: Ensembling large language models with pairwise ranking and generative fusion.
\newblock \emph{arXiv preprint arXiv:2306.02561}.

\bibitem[{Krueger et~al.(2020)Krueger, Leike, Evans, and Salvatier}]{krueger2020active}
Krueger, D., Leike, J., Evans, O., and Salvatier, J. (2020).
\newblock Active reinforcement learning: Observing rewards at a cost.
\newblock \emph{arXiv preprint arXiv:2011.06709}.

\bibitem[{Schulman et~al.(2017)Schulman, Wolski, Dhariwal, Radford, and Klimov}]{schulman2017proximal}
Schulman, J., Wolski, F., Dhariwal, P., Radford, A., and Klimov, O. (2017).
\newblock Proximal policy optimization algorithms.
\newblock \emph{arXiv preprint arXiv:1707.06347}.

\bibitem[{Schulze and Evans(2018)}]{schulze2018active}
Schulze, S. and Evans, O. (2018).
\newblock Active reinforcement learning with monte-carlo tree search.
\newblock \emph{arXiv preprint arXiv:1803.04926}.

\bibitem[{Sekhari et~al.(2023)Sekhari, Sridharan, Sun, and Wu}]{sekhari2023contextual}
Sekhari, A., Sridharan, K., Sun, W., and Wu, R. (2023).
\newblock Contextual bandits and imitation learning with preference-based active queries.
\newblock \emph{Advances in Neural Information Processing Systems}, 36, 11261--11295.

\bibitem[{Shen et~al.(2024)Shen, Zhang, Yao, Zheng, Guo, and Liu}]{shen2024improving}
Shen, W., Zhang, X., Yao, Y., Zheng, R., Guo, H., and Liu, Y. (2024).
\newblock Improving reinforcement learning from human feedback using contrastive rewards.
\newblock \emph{arXiv preprint arXiv:2403.07708}.

\bibitem[{Srinivas et~al.(2012)Srinivas, Krause, Kakade, and Seeger}]{srinivas2012information}
Srinivas, N., Krause, A., Kakade, S.M., and Seeger, M.W. (2012).
\newblock Information-theoretic regret bounds for gaussian process optimization in the bandit setting.
\newblock \emph{IEEE transactions on information theory}, 58(5), 3250--3265.

\bibitem[{Tucker et~al.(2023)Tucker, Biddulph, Wang, and Joachims}]{tucker2023bandits}
Tucker, A.D., Biddulph, C., Wang, C., and Joachims, T. (2023).
\newblock Bandits with costly reward observations.
\newblock In \emph{Uncertainty in Artificial Intelligence}, 2147--2156. PMLR.

\bibitem[{von Werra et~al.(2020)von Werra, Belkada, Tunstall, Beeching, Thrush, Lambert, Huang, Rasul, and Gallouédec}]{vonwerra2022trl}
von Werra, L., Belkada, Y., Tunstall, L., Beeching, E., Thrush, T., Lambert, N., Huang, S., Rasul, K., and Gallouédec, Q. (2020).
\newblock Trl: Transformer reinforcement learning.
\newblock \url{https://github.com/huggingface/trl}.

\bibitem[{Ziegler et~al.(2019)Ziegler, Stiennon, Wu, Brown, Radford, Amodei, Christiano, and Irving}]{ziegler2019fine}
Ziegler, D.M., Stiennon, N., Wu, J., Brown, T.B., Radford, A., Amodei, D., Christiano, P., and Irving, G. (2019).
\newblock Fine-tuning language models from human preferences.
\newblock \emph{arXiv preprint arXiv:1909.08593}.

\end{thebibliography}
\end{document}